\documentclass[runningheads]{llncs}

\usepackage[main=american]{babel} 
\usepackage{amsmath,amssymb,amscd,amsfonts} 
\usepackage{microtype} 
\usepackage[babel,autostyle]{csquotes} 
\usepackage[utf8]{inputenc}
\usepackage[T1]{fontenc}
\usepackage{subcaption}

\usepackage{graphicx}

\newcommand{\EE}{\ensuremath{\mathbb{E}}}

\newcommand{\RR}{\ensuremath{\mathbb{R}}}

\usepackage{verbatim}
\usepackage{array}

\usepackage{tikz}
\usetikzlibrary{bayesnet,matrix}
\tikzset{>=latex}

\usepackage{tikz,tikz-cd}
\usetikzlibrary{cd,patterns,decorations.pathmorphing,decorations.pathreplacing}
\tikzcdset{arrow style=math font}

\usepackage{booktabs}
\usepackage{enumerate} 
\usepackage[lined,boxed]{algorithm2e}
\SetAlCapSkip{5pt}

\usepackage[symbol]{footmisc}

\newcommand{\cate}{{\textsc{cate}}}

\providecommand{\abs}[1]{\ensuremath{\left\lvert#1\right\rvert}}

\DeclareMathOperator*{\argmin}{\arg\min}
\DeclareMathOperator*{\argmax}{\arg\max}

\usepackage{hyperref}
\usepackage{url}

\begin{document}

\title{Neuroevolutionary Feature Representations for Causal
	Inference\thanks{M.B. and G.R. were supported by Adobe Inc. (San Jos\'e,
		Calif., U.S.A.) during the development of this work. A version of
		this manuscript will appear in the proceedings of the 22nd International
		Conference on Computational Science (London, U.K., 21--23 June 2022).}}
\titlerunning{Neuroevolutionary Feature Representations for Causal Inference}
\author{Michael~C.~Burkhart~\inst{1}
	\and Gabriel~Ruiz~\inst{2}
	}
\authorrunning{Burkhart \& Ruiz}
\institute{University of Cambridge, Cambridge, U.K. \\
	\email{mcb93@cam.ac.uk} \and UCLA, Los Angeles, CA, U.S.A. \\
	\email{ruizg@ucla.edu}}
\maketitle

\begin{abstract}
	Within the field of causal inference, we consider the problem of estimating
	heterogeneous treatment effects from data. We propose and validate a novel
	approach for learning feature representations to aid the estimation of the
	conditional average treatment effect or \cate{}. Our method focuses on an
	intermediate layer in a neural network trained to predict the outcome from
	the features. In contrast to previous approaches that encourage the
	distribution of representations to be treatment-invariant, we leverage a
	genetic algorithm that optimizes over representations useful for predicting
	the outcome to select those less useful for predicting the treatment. This
	allows us to retain information within the features useful for predicting
	outcome even if that information may be related to treatment assignment.
	We validate our method on synthetic examples and illustrate its use on a
	real life dataset.
\end{abstract}



\section{Introduction}
\label{s:intro}

In this note, we aim to engineer feature representations to aid in the
estimation of heterogeneous treatment effects. Specifically, we consider the
following graphical model
\begin{equation}
	\label{eq:model}
	\begin{tikzcd}[column sep=small]
		& X \arrow[dl] \arrow[dr] & \\
		W \arrow{rr} & & Y
	\end{tikzcd}
\end{equation}
where $X\in\RR^d$ denotes a vector of features, $W\in\{0,1\}$ represents a
boolean treatment, and $Y\in\RR$ denotes the outcome. Suppose $(X_i, W_i, Y_i)$
for $i=1,\dotsc,n$ are i.i.d. samples from a distribution $P$ respecting the
graph~\eqref{eq:model}. Within the potential outcomes
framework~\cite{Ney23,Rub74}, we let $Y_i(0)$ denote the potential outcome if
$W_i$ were set to $0$ and $Y_i(1)$ denote the potential outcome if $W_i$ were
set to $1$. We wish to estimate the conditional average treatment effect
(\cate{}) defined by
\begin{equation}
	\label{eq:cate}
	\tau(x) = \EE[Y(1)- Y(0)|X=x].
\end{equation}
We impose standard assumptions that the treatment assignment is unconfounded,
meaning that
\[
	\{Y_i(0), Y_i(1)\} \perp W_i \mid X_i,
\]
for all $i$, and random in the sense that
\[
	\epsilon < P(W_i=1 | X_i=x_i) < 1-\epsilon
\]
for all $i$, some $\epsilon>0$, and all $x_i\in\RR^d$ in the support of $X_i$.
These assumptions are jointly known as \emph{strong ignorability}~\cite{Ros83}
and prove sufficient for the \cate{} to be identifiable. Under these
assumptions, there exist well-established methods to estimate the \cate{} from
observed samples (see section~\ref{s:metalearners} for a discussion) that then
allow us to predict the expected individualized impact of an intervention for
novel examples using only their features~\cite{Pea09}. Viewing these approaches
as black box estimators, we aim to learn a mapping $\Phi: \RR^d \to \RR^m$ such
that the estimate of the \cate{} learned from the transformed training data
$(\Phi(X_i), W_i, Y_i)$ is more accurate than an estimate learned on the
original samples $(X_i, W_i, Y_i)$. In particular, we desire a function $\Phi$
yielding a corresponding representation $\Phi(X)$ such that
\begin{enumerate}
	\item $\Phi(X)$ is as useful as $X$ for estimating $Y$, and
	\item among such representations, $\Phi(X)$ is least useful for estimating
	      $W$.
\end{enumerate}
In this way, we hope to produce a new set of features $\Phi(X)$ that retain all
information relevant for predicting the outcome, but are less related to
treatment assignment. \emph{We propose learning $\Phi$ as an intermediate layer
	in a neural network estimating a functional relationship of $Y$ given $X$.
	We apply a genetic algorithm~\cite{Hol75} to a population of such mappings
	to evolve and select the one for which the associated representation $\Phi(X)$
	is least useful for approximating $W$.}

Feature representations are commonly used in machine learning to aid in the
training of supervised models~\cite{Ben13} and have been previously
demonstrated to aid in causal modeling. Johansson, et al.~\cite{Joh16,Sha17b}
viewed counterfactual inference on observational data as a covariate shift
problem and learned neural network-based representations designed to produce
similar empirical distributions among the treatment and control populations,
namely $\{\Phi(X_i)\}_{W_i=1}$ and $\{\Phi(X_i)\}_{W_i=0}$. Li \&
Fu~\cite{Li17a} and Yao, et al.~\cite{Yao18} developed representations in a
related vein designed to preserve local similarity. We generally agree with
Zhang et al.'s~\cite{Zha20} recent argument that domain invariance
often removes too much information from the features for causal
inference.\footnote[2]{Zhao et al.~\cite{Zha19a} make this argument in a more
	general setting.} \emph{In contrast to most previous approaches, we develop a
	feature representation that attempts to preserve information useful for
	predicting the treatment effect if it is also useful for predicting the
	outcome.}

\subsubsection{Outline.}
We proceed as follows. In the next section, we describe methods for learning
the \cate{} from observational data and introduce genetic algorithms. In
section~\ref{s:methodology}, we describe our methodology in full. We then
validate our method on artificial data in section~\ref{s:ablation_synthetic}
and on a publicly available experimental dataset in
section~\ref{s:in_house_data}, before concluding in
section~\ref{s:conclusions}.

\section{Related work}
\label{s:related}

In the first part of this section, we discuss standard methods for learning the
\cate{} function from data. We will subsequently use these to test our proposed
feature engineering methods in section~\ref{s:ablation_synthetic}. In the
second part, we briefly outline evolutionary algorithms for training neural
networks, commonly called neuroevolutionary methods.

\subsection{Meta-learners}
\label{s:metalearners}
We adopt the standard assumptions of unconfoundedness and the random assignment
of treatment effects that together constitute strong ignorability. Given i.i.d.
samples from a distribution $P$ respecting~\eqref{eq:model} and these
assumptions, there exist numerous meta-learning approaches that leverage an
arbitrary regression framework (e.g., random forests, neural networks,
linear regression models) to estimate the \cate{} that we now describe.

\subsubsection*{S-learner.}
The S-learner (single-learner) uses a standard supervised learner
(regression model) to estimate $\mu(x,w) = \EE[Y | X=x, W=w]$ from observation
data and then predicts
\[
	\hat\tau_S(x) = \hat\mu(x,1) - \hat\mu(x,0)
\]
where we use the standard hat notation to denote estimated versions of the
underlying functions.

\subsubsection*{T-learner.}
The T-learner (two-learner) estimates $\mu_1(x) = \EE[Y(1)|X=x]$ from observed
treatment data $\{(X_i,Y_i)\}_{W_i=1}$ and $\mu_0(x) = \EE[Y(0)|X=x]$ from
observed control data $\{(X_i,Y_i)\}_{W_i=0}$, and then predicts
\[
	\hat\tau_T(x) = \hat\mu_1(x) - \hat\mu_0(x).
\]

\subsubsection*{X-learner.}
The X-learner~\cite{Kun19} estimates $\mu_1$ and $\mu_0$ as in the T-learner,
and then predicts the contrapositive outcome for each training point. Next,
the algorithm estimates $\tau_1(x)=\EE[\tilde D_i^1 \mid X=x]$ on
$\{(X_i,\tilde D_i^1) \}_{W_i=1}$ where $\tilde D_i^1 = Y_i - \hat\mu_0(X_i)$
and $\tau_0(x)=\EE[\tilde D_i^0 \mid X=x]$ on $\{(X_i,\tilde D_i^0)
	\}_{W_i=0}$ where $\tilde D_i^0 = \hat\mu_1(X_i)-Y_i$. The X-learner then
predicts
\[
	\hat\tau_X(x) = g(x)\hat\tau_0(x) + (1-g(x))\hat\tau_1(x)
\]
where $g: \RR^d\to [0,1]$ is a weight function. The creators of the X-learner
remark that the treatment propensity function~\eqref{eq:treatment_propensity}
often works well for $g$, as do the constant functions $1$ and $0$. In our
implementation, we use $g(x)\equiv 1/2$.

Concerning this method, we note that it is also possible to directly estimate
$\tau$ from
\[
	\{(X_i,Y_i - \hat\mu_0(X_i)) \}_{W_i=1}
	\cup \{(X_i,\hat\mu_1(X_i)-Y_i) \}_{W_i=0}
\]
or, using $\hat\mu(x,w)$ from the S-learner approach, with
\[
	\{(X_i,Y_i - \hat\mu(X_i,0)) \}_{W_i=1}
	\cup \{(X_i,\hat\mu(X_i,1)-Y_i) \}_{W_i=0}.
\]
We find that these alternate approaches work well in practice and obviate the
need to estimate or fix $g$.

\subsubsection*{R-learner.}
Within the setting of the graphical model~\eqref{eq:model}, we define the
treatment propensity (sometimes called the propensity score) as
\begin{equation}
	\label{eq:treatment_propensity}
	e(x) = P(W=1 | X=x)
\end{equation}
and the conditional mean outcome as
\begin{equation}
	\label{eq:conditional_mean_outcome}
	m(x) = \EE[Y | X=x].
\end{equation}
The R-learner~\cite{Nie21} leverages Robinson's decomposition~\cite{Rob88} that
led to Robin's reformulation~\cite{Rob04} of the \cate{} function as the
solution to the optimization problem
\begin{equation}
	\label{eq:cate_as_optimization}
	\tau(\cdot) = \argmin_{\tau} \bigg\{ \EE_{(X,W,Y)\sim P}
	\bigg[ \abs{\big(Y - m(X)\big) - \big(W - e(X)\big)\tau(X)}^2 \bigg] \bigg\}
\end{equation}
in terms of the treatment propensity~\eqref{eq:treatment_propensity} and the
conditional mean outcome~\eqref{eq:conditional_mean_outcome}. In practice, a
regularized, empirical version of~\eqref{eq:cate_as_optimization} is minimized
via a two-step process: (1) cross-validated estimates $\hat m$ and $\hat e$ are
obtained for $m$ and $e$, respectively, and then (2) the empirical loss is
evaluated using folds of the data not used for estimating $\hat m$ and $\hat
	e$, and then minimized. The authors Nie \& Wager note that the structure of the
loss function eliminates correlations between $m$ and $e$ while allowing one to
separately specify the form of $\tau$ through the choice of optimization
method. In this paper, the only R-learner we use is the causal forest as
implemented with generalized random forests~\cite{Ath19} using the default
options, including honest splitting~\cite{Wag18}.

\subsection{Genetic and neuroevolutionary algorithms}
Holland introduced genetic algorithms~\cite{Hol75} as a nature-inspired
approach to optimization. Generally speaking, these algorithms produce
successive generations of candidate solutions. New generations are formed by
selecting the fittest members from the previous generation and performing
cross-over and/or mutation operations on them to produce new offspring
candidates. Evolutionary algorithms encompass extensions and generalizations to
this approach including memetic algorithms~\cite{Mos89} that perform local
refinements, genetic programming~\cite{For81} that acts on programs represented
as trees, and evolutionary programming~\cite{Fog62} and
strategies~\cite{Rec70,Sch75} that operate on more general representations.
When such methods are applied specifically to the design and training of neural
networks, they are commonly known as neuroevolutionary algorithms. See Stanley
et al.~\cite{Sta19} for a comprehensive survey. In the next section, we
describe a specific neuroevolutionary strategy for feature engineering.

\section{Methodology}
\label{s:methodology}

In this section, we describe how we form our feature mapping
$\Phi:\RR^d\to\RR^m$. To generate a single candidate solution, we train a
shallow neural network to predict $Y$ from $X$ and extract an intermediate
layer of this network. Each candidate map created in this way should yield a
representation as functionally useful for predicting $Y$ as $X$ is. We then
iteratively evolve cohorts of parameter sets for such maps to create a
representation that carries the least amount of useful information for
predicting the treatment $W$.

\subsection{Candidate solutions}
We consider neural networks $f_\Theta: \RR^d \to \RR$ of the form
\begin{equation}
	\label{eq:f_theta}
	f_\Theta(x) = M_2 \cdot a( M_1 \cdot x + b_1 ) + b_2
\end{equation}
where $M_1\in \RR^{m \times d}$, $M_2 \in \RR^{1\times m}$ are real-valued
matrices (often called weights), $b_1\in \RR^m$ and $b_2 \in \RR^1$ are vectors
(often called biases), and $a$ is a nonlinear activation function applied
component-wise. We let $\Theta = ( M_1, M_2, b_1, b_2)$ denote the parameters
for $f_\Theta$. Though $f_\Theta$ is decidedly not a deep neural network, we
note that, as a neural network with a single hidden layer, it remains a
universal function approximator in the sense of Hornik et al.~\cite{Hor89}.
Optimizing the network~\eqref{eq:f_theta} in order to best predict $Y$ from $X$
seeks the solution
\begin{equation}
	\label{eq:optimize_f}
	\Theta_* = \argmin_\Theta \EE \abs{ Y - f_\Theta(X) }^2.
\end{equation}
Given parameters $\Theta$ for the network~\eqref{eq:f_theta}, we let
$\Phi_\Theta: \RR^d\to\RR^m$ given by
\begin{equation}
	\label{eq:phi_theta}
	\Phi_\Theta(x) = a( M_1 \cdot x + b_1 )
\end{equation}
denote the output of the hidden layer. We restrict to candidate feature
mappings of this form. As these mappings are completely characterized by their
associated parameters, we define a fitness function and evolutionary algorithm
directly in terms of parameter sets $\Theta$ in the following subsections.

\subsection{Fitness function}
\begin{figure}[htb]
	\centering
	\begin{minipage}[b]{0.4\textwidth}
		\centering
		\includegraphics[width=\textwidth]{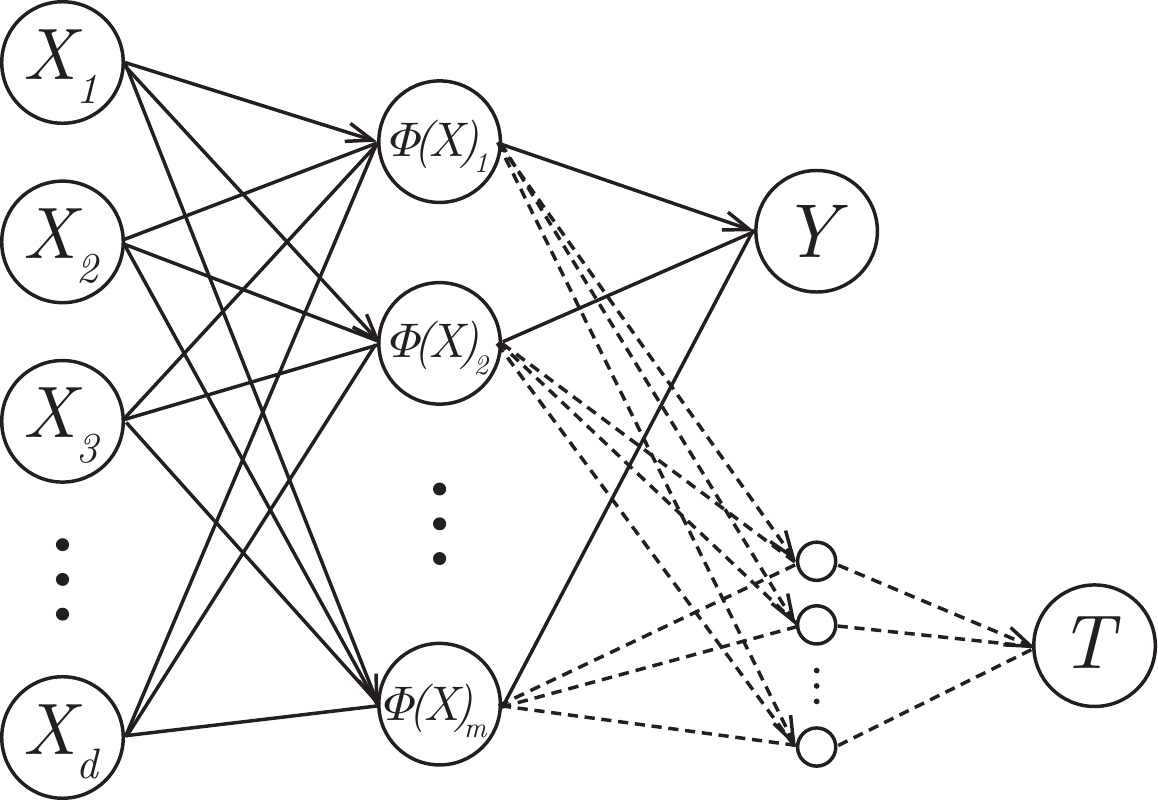}
	\end{minipage}
	\hfill
	\begin{minipage}[b]{0.55\textwidth}
		\caption{In this schematic, the arrows connecting the $X_i$ to the
			$\Phi(X)_j$ represent the map $\Phi_\Theta$ as
			in~\eqref{eq:phi_theta}; these, in addition to the ones joining the
			$\Phi(X)_j$ to $Y$, represent $f_\Theta$ given explicitly
			in~\eqref{eq:f_theta}; and the original arrows along with the
			dashed arrows represent $g_{\Psi,\Theta}$ as in~\eqref{eq:g_psi}.}
		\label{f:schematic}
	\end{minipage}
\end{figure}

For parameters $\Theta$ near the optimum~\eqref{eq:optimize_f}, we note that
$\Phi_\Theta(X)$ should be approximately as useful as $X$ for learning a
functional relationship with $Y$. However, for some values of $\Theta$, the
mapped features $\Phi_\Theta(X)$ may carry information useful for predicting
$W$, and for this reason we consider a network $g_{\Psi,\Theta}: \RR^d\to[0,1]$
given by
\begin{equation}
	\label{eq:g_psi}
	g_{\Psi,\Theta}(x)
	= \sigma( M_4 \cdot a (M_3 \cdot \Phi_\Theta(x) + b_3) + b_4)
\end{equation}
where $M_3 \in \RR^{k \times m}$ and $M_4 \in \RR^{1 \times k}$ are weights,
$b_3\in \RR^k$ and $b_4 \in \RR$ are biases, $a$ is a nonlinear activation
function applied component-wise,\footnote{though not necessarily the same as
	the one used in~\eqref{eq:f_theta} and \eqref{eq:phi_theta}} and $\sigma(x) =
	(1+\exp(-x))^{-1}$ denotes the sigmoidal activation function. In this case,
$\Psi = ( M_3, M_4, b_3, b_4)$ denotes the collection of tunable parameters. We
define the fitness of a parameter set $\Theta$ to be
\begin{equation}
	\label{eq:mu_theta}
	\mu(\Theta) = \min_{\Psi} \EE \abs{ W - g_{\Psi,\Theta}(X)}^2.
\end{equation}
In this way, we express a preference for representations $\Phi_\Theta(X)$ that
are less useful for predicting $W$. For a schematic of these architectures,
please see \figurename~\ref{f:schematic}.

\subsection{Evolutionary algorithm}
We now describe a method to generate and evolve a cohort of candidate parameter
sets $\Theta$ intended to seek a parameter set $\Theta_*$ such that
$\Phi_{\Theta*}(X)$ is nearly as useful for predicting $Y$ as $X$ is and, among
such representations, $\Phi_{\Theta*}(X)$ is least useful for predicting $W$.

Given training data $(X_i, W_i, Y_i) \sim^\text{i.i.d.} P$ for $1\leq i \leq
	n$, we first partition the data into training and validation sets. We form an
initial cohort of $c$ candidates independently as follows. For $1\leq j \leq
	c$, we randomly instantiate $\Theta_j$ using Glorot normal
initialization~\cite{Glo10} for the weights and zeros for the biases and then
apply batch-based gradient descent on the training set to seek the
solution to~\eqref{eq:optimize_f}. In particular, we use the Adam
optimizer~\cite{Kin15} that maintains parameter-specific learning
rates~\cite[cf. AdaGrad]{Duc11} and allows these rates to sometimes
increase~\cite[cf. Adadelta]{Zei12} by adapting them using the first two
moments from recent gradient updates. We use Tikhonov
regularization~\cite{Tik63} for the weights and apply a dropout
layer~\cite{Sri14} after the $a(x) = \tanh(x)$ activation function\footnote[4]{
	We tested rectified~\cite{Nai10} and exponential~\cite{Cle16} linear unit
	activation functions for $a$ in $\Phi_\Theta$ but noticed only minor
	differences in subsequent performance of the causal forest.}%
to prevent overfitting.

For each constituent $\Theta_j$ in the cohort, we then initialize and train a
network $g_{\Psi,\Theta_j}$ as in~\eqref{eq:g_psi} to seek $\Psi_j =
	\argmin_{\Psi} \EE \abs{ W - g_{\Psi,\Theta_j}(X)}^2$ on the training set and
then evaluate $\EE \abs{ W - g_{\Psi_j,\Theta_j}(X)}^2$ empirically on the
validation set to estimate $\mu(\Theta_j)$. We then use the $\ell$ fittest
members of the current cohort to form a new cohort as follows. For each of the
$\binom{\ell}{2}$ pairings, we apply Montana and Davis's node-based
crossover~\cite{Mon89} method to the parameters $M_1$ and $b_1$ that we use to
form $\Phi$. This amounts to forming a new $\Phi$ by randomly selecting one of
the two parents and using that parent's mapping for each coordinate. Thus, the
new $M_1$ and $b_1$ are selected in a row-wise manner from the corresponding
rows of the parents, and then the new $M_2$ and $b_2$ are randomly initialized
and a few steps of optimization are performed to form the offspring candidate.
The next generation then consists of the best performing candidate from the
previous generation, $\binom{\ell}{2}$~candidates formed by crossing the best
$\ell$ candidates of the previous generation, and $c-1-\binom{\ell}{2}$
entirely new candidates generated from scratch.

\begin{algorithm}[!tb]
	\SetAlgoLined
	\KwData{training $\mathcal{T} = \{(X_i, W_i, Y_i)\}_{i \in T}$ and
	validation $\mathcal{V} = \{(X_i, W_i, Y_i)\}_{i \in V}$ datasets drawn
	i.i.d. from $P$ respecting the graphical model~\eqref{eq:model}\; positive
	integer parameters: $c$ cohort size, $\ell$ number of members involved in
	forming the next generation, $g$ number of generations, $m$ dimensionality
	of the latent representation, and $k$ dimensionality of the final hidden
	layer in~\eqref{eq:g_psi}}
	\KwResult{parameterized function $\Phi_{\Theta_*}: \RR^d \to \RR^m$ such
	that models for the \cate{} learned using the transformed training data
	$\{(\Phi(X_i), W_i, Y_i)\}_{i\in \mathcal{T}}$ perform better than those
	learned on the original dataset $\mathcal{T}$}
	\For{$j = 1,\dotsc,c$}{ Optimize a parameter set $\Theta_j$ to seek
		\eqref{eq:optimize_f} on training batches from a random initialization\;
	}
	Form the first generation $\mathcal G_1 = \{ \Theta_j \}_{j = 1,\dotsc, c}$\;
	\For(\#\emph{form the next generation}){$t = 2,\dotsc,g$}{ Initialize
		new generation $\mathcal G_t = \{ \argmax_{\Theta\in\mathcal{G}_{t-1}}
			\mu(\Theta) \}$ with the best-performing candidate from the previous
		generation\;
		\For(\#\emph{form new candidates using crossover}){unique pairs
			$\{\Theta_j, \Theta_k\}$ formed from the top $\ell$ candidates from
			$\mathcal{G}_{t-1}$}{
			Initialize $M_1\in \RR^{m\times d}$ and $b_1\in\RR^m$ as the $M_1$ and
			$b_1$ from $\Theta_j$\; \For{$ \kappa= 1,\dotsc,m$}{ Let $\xi\sim
					\operatorname{Bernoulli}(1/2)$\; \lIf{$\xi = 1$}{
					replace the $\kappa$th row of $M_1$ and the $\kappa$th component of
					$b_1$ with those from $\Theta_k$}
			}
			Randomly initialize $M_2$ and $b_2$ and take optimization steps towards
			the solution to~\eqref{eq:optimize_f}\; Add $\Theta = ( M_1, M_2, b_1,
				b_2)$ to $\mathcal G_t$
		}
		\While{$\abs{\mathcal G_t}<c$}{ Optimize a parameter set $\Theta$ to seek
			\eqref{eq:optimize_f} on training batches from a random initialization\;
			Add $\Theta$ to $\mathcal G_t$
		}
	}
	Let $\Theta_* = \argmax_{\Theta\in\mathcal{G}_g} \mu(\Theta)$\;
	\Return{$\Phi_{\Theta_*}$ as in \eqref{eq:phi_theta}}
	\caption{Neuroevolutionary Feature Engineering for Causal Inference}
	\label{algo:main}
\end{algorithm}

We summarize our approach using pseudo-code in
\algorithmcfname~\ref{algo:main}. All computation of the valuation function is
done by training a network of the form~\eqref{eq:g_psi} to minimize $\abs{ W -
		g_{\Psi,\Theta}(X)}^2$ on training batches and then
approximating~\eqref{eq:mu_theta} by taking the empirical mean on the
validation set.

\subsection{Remark on linearity}
Due to our choice of representation $\Phi$ in~\eqref{eq:phi_theta}, after
training the network~\eqref{eq:f_theta} to optimize~\eqref{eq:optimize_f}, we
expect the relationship between the learned features $\Phi(X)$ and the outcome
$Y$ to be approximately linear. In particular, we will have $Y\approx M_2\cdot
	\Phi(X)$ for $M_2$ as given in~\eqref{eq:f_theta}. For this reason, the
causal meta-learners trained using a linear regression base learner may
benefit more extensively from using the transformed features instead of the
original features, especially in cases where the relationship between the
original features and outcomes is not well-approximated as linear. We provide
specific examples in the next section involving meta-learners trained with
ridge regression.

\subsection{Remark on our assumptions}
In order to use the represented features $\Phi(X_i)$ in place of the original
features $X_i$ when learning the \cate{}, we require that strong ignorability
holds for the transformed dataset $(\Phi(X_i), W_i, Y_i)$, $i=1,\dotsc,n$. One
sufficient, though generally not necessary, assumption that would imply strong
ignorability is for $\Phi$ to be invertible on the support of
$X$~\cite[assumption 1]{Sha17b}. Unconfoundedness would also be guaranteed if
$\Phi(X)$ satisfied the backdoor condition with respect to
$(W,Y)$~\cite[section 3.3.1]{Pea09}.

\section{Ablation study on generated data}
\label{s:ablation_synthetic}

Due to the fundamental challenge of causal inference (namely, that the
counterfactual outcome cannot be observed, even in controlled experiments), it
is common practice to compare approaches to \cate{} estimation on artificially
generated datasets that allow the \cate{} to be calculated directly for
evaluative purposes. In this section, we perform experiments using two such
data generation mechanisms from Nie \& Wager's paper~\cite{Nie21} that we now
describe.\footnote[3]{Nie \& Wager's paper included four setups,
	namely A--D; however setup B modeled a controlled randomized trial and setup D
	had unrelated treatment and control arms.} Both setups provide a joint
distribution satisfying the graph~\eqref{eq:model}. For the vector-valued
random variable $X\in\RR^d$, we let $X_{ij}$ denote $j$th component ($1\leq j
	\leq d$) of the $i$th sample ($1\leq i \leq n$). The specifics for both setups
are given as follows.

\subsubsection{Setup A.}
For $\sigma>0$ and an integer $d>0$, we let
\[
	X_i\sim^\text{i.i.d.} \operatorname{Uniform}([0,1]^d)
	\quad\text{and}\quad
	W_i \mid X_i \sim \operatorname{Bernoulli}(e(X_i)),
\]
where $e(X_i) = \max\{0.1, \min\{\sin(\pi X_{i1}X_{i2}), 0.9\} \}$ and
\[
	Y_i \mid X_i,W_i
	\sim \mathcal{N}\big(b(X_i) + (W_i-0.5)\tau(X_i),\sigma^2\big),
\]
where $b(X_i) = \sin(\pi X_{i1}X_{i2}) + 2 (X_{i3}-0.5)^2 + X_{i4} + 0.5
	X_{i5}$ and $\tau(X_i) = (X_{i1}+X_{i2}) /2$. In this paper, we let $d=24$,
$n=200$, and $\sigma=1$.

\subsubsection{Setup C.}
For $\sigma>0$ and an integer $d>0$, we let
\[
	X_i\sim^\text{i.i.d.} \mathcal{N}_d(\vec 0, I_{d\times d})
	\quad\text{and}\quad
	W_i \mid X_i \sim \operatorname{Bernoulli}(e(X_i)),
\]
where in this case $e(X_i) = (1+ \exp(X_{i2} + X_{i3}) )^{-1}$ and
\[
	Y_i \mid X_i, W_i
	\sim \mathcal{N}\big( b(X_i) + (W_i-0.5) \tau(X_i), \sigma^2\big),
\]
where now $b(X_i) = 2 \log(1+ \exp(X_{i1} + X_{i2} + X_{i3}))$
and $\tau(X_i) = 1$. In our example, we let $d=12$, $n=500$, and $\sigma=1$.

\subsection{Comparison methodology}
For each data generation method, we ran 100 independent trials. Within each
trial, we simulated a dataset of size $n$ and randomly partitioned it into
training, validation, and testing subsets at a 70\%-15\%-15\% rate. We trained
causal inference methods on the training set, using the validation data to aid
the training of some base estimators for the meta-learners, and predicted on
the test dataset. We then developed a feature map using the training and
validation data as described in the previous section, applied this map to all
features, and repeated the training and testing process using the new features.

To determine the impact of the fitness selection process, we also learned a
feature transformation that did not make use of the fitness function at all. In
effect, it simply generated a single candidate mapping and used it to transform
all the features (without any cross-over or further mutation). Features
developed in this way are described as ``no fitness'' in the tables that
follow.

We compared the causal forest with default options (as found in R's grf
package), and the S-, T-, and X-learners as described in
section~\ref{s:metalearners} with two base learners. The first base learner is
LightGBM~\cite{Ke17}, a boosted random forest algorithm that introduced novel
techniques for sampling and feature bundling. The second is a cross-validated
ridge regression model (as found in scikit-learn) that performs multiple linear
regression with an $L^2$ normalization on the weights.

\subsection{Results}

We report results for setup A in \tablename~\ref{t:setup_A_results} and results
for setup C in \tablename~\ref{t:setup_C_results}. For both setups, we
consider a paired t-test for equal means against a two-sided alternative (as
implemented in Python's scipy package). For setup A, we find that the
improvement in MSE from using the transformed features in place of the original
features corresponds to a statistically significant difference for the
following learners: the causal forest ($p<0.001$), the S-learner with ridge
regression ($p<0.001$), the T-learner with both LightGBM ($p<0.001$) and ridge
regression ($p<0.001$), and the X-learner with both LightGBM ($p<0.001$) and
ridge regression ($p<0.001$). For setup C, we again find significant
differences for the causal forest ($p=0.023$), S-learner with LightGBM
($p=0.003$), T-learner with ridge regression ($p<0.001$) and X-learner with
ridge regression ($p<0.001$).

In summary, we find that our feature transformation method improves the
performance of multiple standard estimators for the \cate{} under two data
generation models.

\begin{table}[htb]
	\centering
	\begin{tabular}{llccccccc>{\bfseries}c}
		\toprule
		learner       & features    & tr.\,1 & tr.\,2 & tr.\,3 & \dots & tr.\,98 & tr.\,99 & tr.\,100 & avg.  \\ \midrule
		Causal forest & initial     & 0.488  & 0.161  & 0.042  & \dots & 0.207   & 0.212   & 0.592    & 0.175 \\
		              & no fitness  & 0.305  & 0.050  & 0.035  & \dots & 0.048   & 0.068   & 0.353    & 0.114 \\
		              & transformed & 0.163  & 0.054  & 0.044  & \dots & 0.046   & 0.103   & 0.386    & 0.120 \\ \midrule
		S-L. w/ LGBM  & initial     & 0.121  & 0.055  & 0.191  & \dots & 0.048   & 0.059   & 0.323    & 0.149 \\
		              & no fitness  & 0.171  & 0.111  & 0.089  & \dots & 0.149   & 0.098   & 0.395    & 0.140 \\
		              & transformed & 0.157  & 0.077  & 0.096  & \dots & 0.168   & 0.176   & 0.241    & 0.135 \\ \midrule
		S-L. w/ Ridge & initial     & 0.257  & 0.038  & 0.032  & \dots & 0.048   & 0.067   & 0.290    & 0.093 \\
		              & no fitness  & 0.202  & 0.038  & 0.037  & \dots & 0.039   & 0.046   & 0.249    & 0.079 \\
		              & transformed & 0.110  & 0.039  & 0.051  & \dots & 0.045   & 0.041   & 0.274    & 0.081 \\ \midrule
		T-L. w/ LGBM  & initial     & 0.574  & 0.351  & 0.691  & \dots & 0.448   & 0.198   & 1.189    & 0.666 \\
		              & no fitness  & 0.647  & 0.516  & 0.596  & \dots & 0.393   & 0.303   & 0.669    & 0.536 \\
		              & transformed & 0.469  & 0.295  & 0.599  & \dots & 0.334   & 0.504   & 1.056    & 0.512 \\ \midrule
		T-L. w/ Ridge & initial     & 0.763  & 0.254  & 0.888  & \dots & 0.916   & 0.260   & 0.553    & 0.745 \\
		              & no fitness  & 0.257  & 0.177  & 0.103  & \dots & 0.237   & 0.114   & 0.574    & 0.333 \\
		              & transformed & 0.147  & 0.042  & 0.206  & \dots & 0.087   & 0.118   & 0.471    & 0.325 \\ \midrule
		X-L. w/ LGBM  & initial     & 0.339  & 0.207  & 0.309  & \dots & 0.394   & 0.134   & 0.811    & 0.411 \\
		              & no fitness  & 0.497  & 0.225  & 0.230  & \dots & 0.161   & 0.128   & 0.493    & 0.335 \\
		              & transformed & 0.361  & 0.135  & 0.317  & \dots & 0.124   & 0.368   & 0.757    & 0.317 \\ \midrule
		X-L. w/ Ridge & initial     & 0.643  & 0.221  & 0.702  & \dots & 0.798   & 0.240   & 0.470    & 0.630 \\
		              & no fitness  & 0.247  & 0.145  & 0.097  & \dots & 0.197   & 0.110   & 0.540    & 0.289 \\
		              & transformed & 0.149  & 0.033  & 0.176  & \dots & 0.066   & 0.118   & 0.460    & 0.288 \\
		\bottomrule
	\end{tabular}
	\vspace{4pt}
	\caption{Mean Squared Error (MSE) over 100 independent trials run using
		setup~A. Algorithm~\ref{algo:main} was run with parameters: cohort size
		$c=4$, progenitors $\ell=2$, number of cohorts $g=5$, representation
		dimensionality $m=20$, and fitness function parameter $k=10$.}
	\label{t:setup_A_results}
\end{table}

\begin{table}[htb]
	\centering
	\begin{tabular}{llccccccc>{\bfseries}c}
		\toprule
		learner       & features    & tr.\,1 & tr.\,2 & tr.\,3 & \dots & tr.\,98 & tr.\,99 & tr.\,100 & avg.  \\ \midrule
		Causal forest & initial     & 0.002  & 0.062  & 0.016  & \dots & 0.023   & 0.008   & 0.071    & 0.035 \\
		              & no fitness  & 0.004  & 0.027  & 0.028  & \dots & 0.046   & 0.006   & 0.047    & 0.029 \\
		              & transformed & 0.007  & 0.024  & 0.028  & \dots & 0.045   & 0.003   & 0.034    & 0.029 \\ \midrule
		S-L. w/ LGBM  & initial     & 0.247  & 0.244  & 0.279  & \dots & 0.222   & 0.190   & 0.202    & 0.226 \\
		              & no fitness  & 0.173  & 0.207  & 0.207  & \dots & 0.214   & 0.164   & 0.198    & 0.211 \\
		              & transformed & 0.188  & 0.183  & 0.146  & \dots & 0.331   & 0.149   & 0.196    & 0.204 \\ \midrule
		S-L. w/ Ridge & initial     & 0.000  & 0.024  & 0.017  & \dots & 0.004   & 0.012   & 0.005    & 0.015 \\
		              & no fitness  & 0.000  & 0.016  & 0.017  & \dots & 0.005   & 0.012   & 0.005    & 0.015 \\
		              & transformed & 0.000  & 0.022  & 0.018  & \dots & 0.005   & 0.025   & 0.006    & 0.015 \\ \midrule
		T-L. w/ LGBM  & initial     & 0.771  & 0.482  & 0.833  & \dots & 0.490   & 0.538   & 0.424    & 0.567 \\
		              & no fitness  & 0.587  & 0.543  & 0.593  & \dots & 0.491   & 0.766   & 0.537    & 0.551 \\
		              & transformed & 0.353  & 0.541  & 0.378  & \dots & 0.551   & 0.575   & 0.531    & 0.544 \\ \midrule
		T-L. w/ Ridge & initial     & 0.143  & 0.260  & 0.189  & \dots & 0.173   & 0.148   & 0.124    & 0.178 \\
		              & no fitness  & 0.049  & 0.164  & 0.096  & \dots & 0.145   & 0.106   & 0.093    & 0.125 \\
		              & transformed & 0.084  & 0.136  & 0.089  & \dots & 0.116   & 0.095   & 0.109    & 0.128 \\ \midrule
		X-L. w/ LGBM  & initial     & 0.286  & 0.216  & 0.474  & \dots & 0.282   & 0.247   & 0.275    & 0.313 \\
		              & no fitness  & 0.348  & 0.236  & 0.361  & \dots & 0.318   & 0.377   & 0.254    & 0.313 \\
		              & transformed & 0.342  & 0.193  & 0.207  & \dots & 0.208   & 0.278   & 0.231    & 0.298 \\ \midrule
		X-L. w/ Ridge & initial     & 0.129  & 0.241  & 0.170  & \dots & 0.156   & 0.136   & 0.114    & 0.166 \\
		              & no fitness  & 0.039  & 0.142  & 0.081  & \dots & 0.119   & 0.099   & 0.079    & 0.109 \\
		              & transformed & 0.072  & 0.124  & 0.078  & \dots & 0.104   & 0.082   & 0.097    & 0.114 \\
		\bottomrule
	\end{tabular}
	\vspace{4pt}
	\caption{Mean Squared Error (MSE) over 100 independent trials run using
		setup~C. Algorithm~\ref{algo:main} was run with
		$c=4$, $\ell=2$, $g=2$, $m=20$, and $k=10$.}
	\label{t:setup_C_results}
\end{table}

\section{Application to econometric data}
\label{s:in_house_data}

In this section, we apply our feature engineering method to the LaLonde
dataset~\cite{LaL86,Deh99} chronicling the results of an experimental study on
temporary employment opportunities. The dataset contains information from 445
participants who were randomly assigned to either an experimental group that
received a temporary job and career counseling or to a control group that
received no assistance. Features include age and education (in years), earnings
in 1974 (in \$, prior to treatment), and indicators for African-American
heritage, Hispanic-American heritage, marital status, and possession of a high
school diploma. We consider the outcome of earnings in 1978 (in \$, after
treatment).

\begin{figure}[tb]
	\centering
	\begin{subfigure}[b]{0.475\textwidth}
		\centering
		\includegraphics[width=\textwidth]{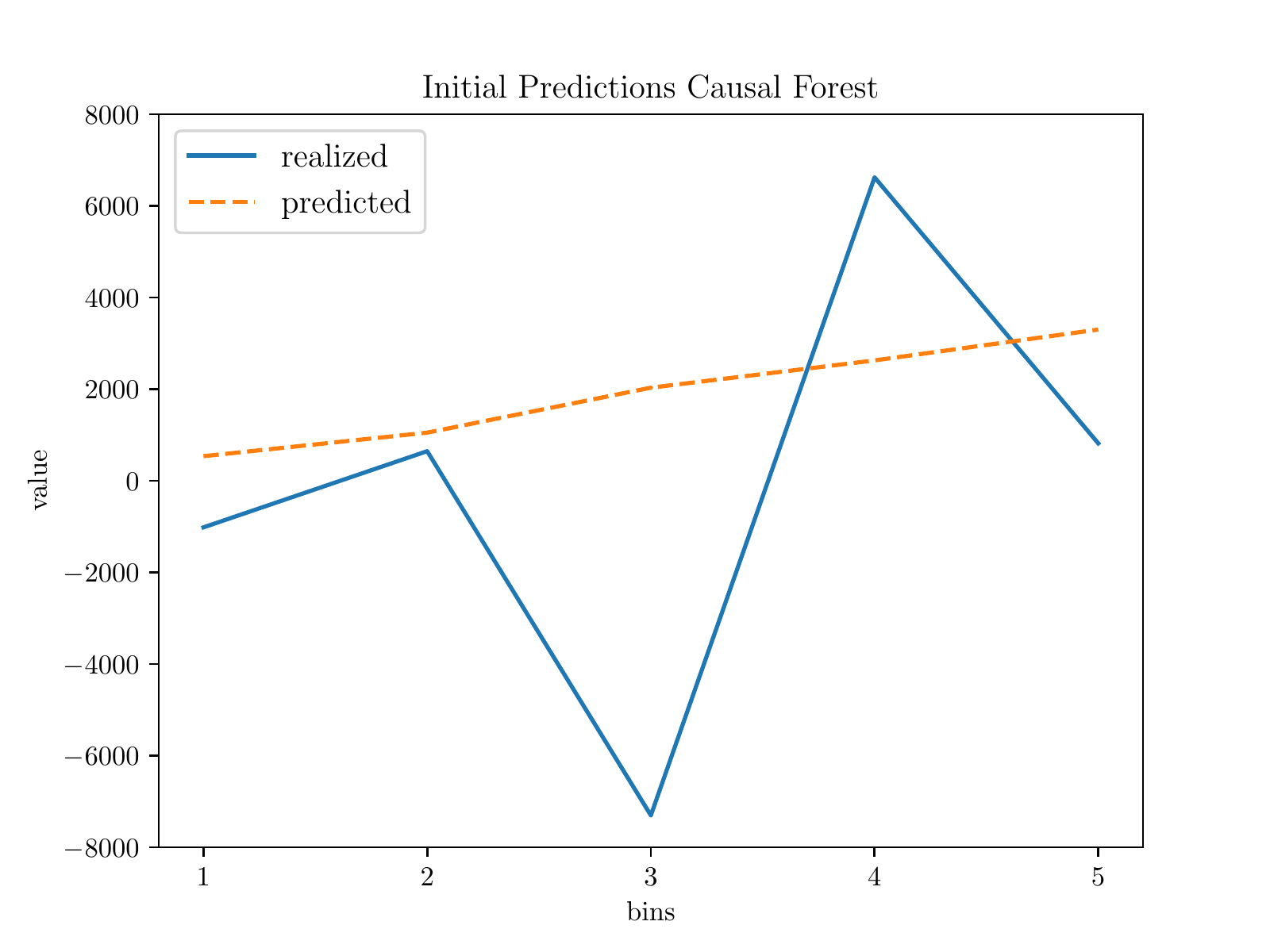}
	\end{subfigure}
	\hspace{10pt}
	\begin{subfigure}[b]{0.475\textwidth}
		\centering
		\includegraphics[width=\textwidth]{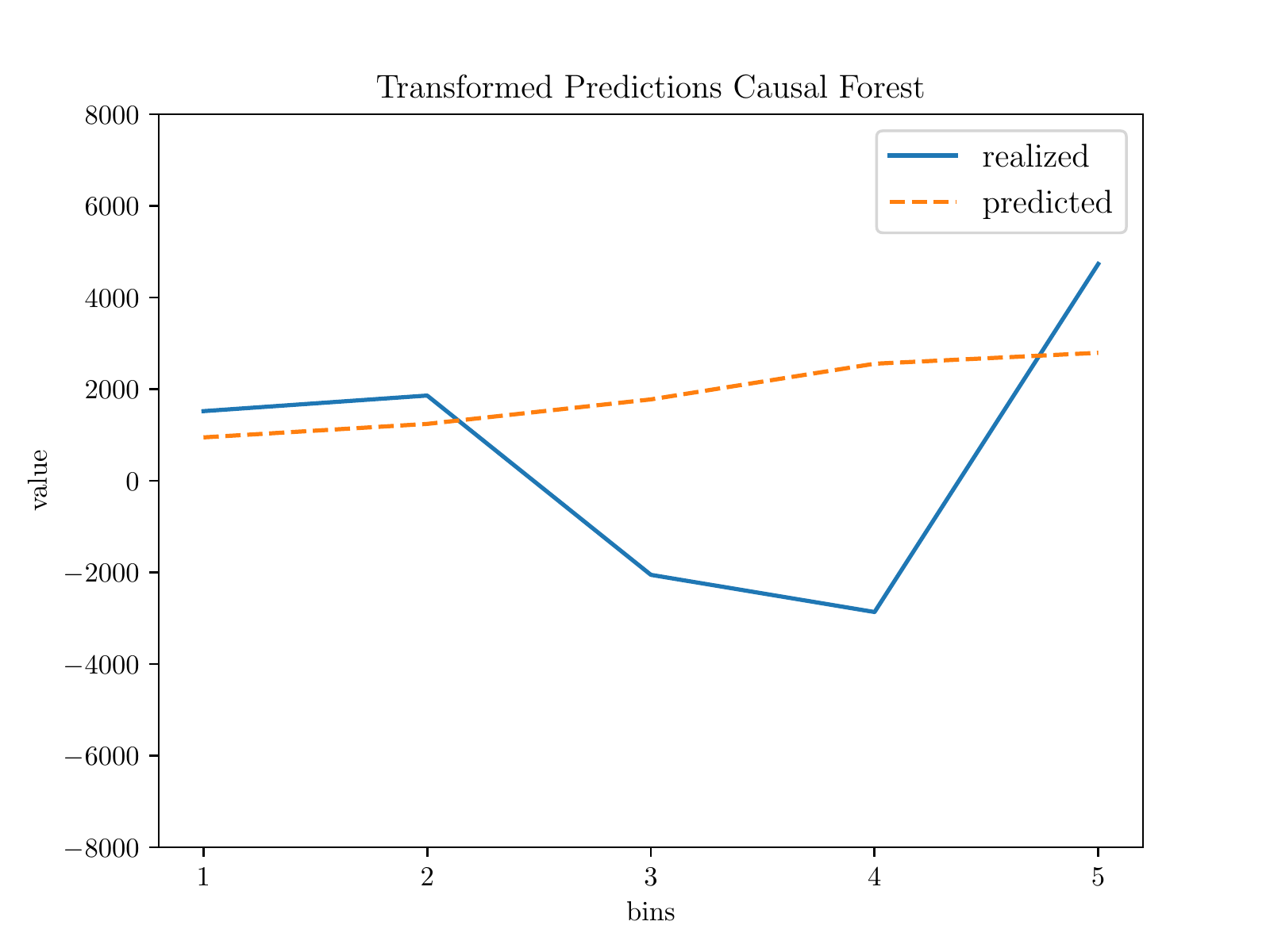}
	\end{subfigure}
	\caption{We plot estimated realized and predicted average treatment effects
		versus the quintiles of predicted treatment effect (5 bins) for a
		causal forest using (left) the initial features and (right) the features
		transformed using our method. In order to estimate realized treatment
		effect within each bin, we take the difference between the average
		outcome for persons randomly assigned to the experimental group and the
		average outcome for persons randomly assigned to the control group.}
	\label{fig:lalonde}
\end{figure}

We cannot determine true average treatment effects based on individual-level
characteristics (i.e. the true \cate{} values) for real life experimental data
as we can with the synthetic examples of the previous section. Instead, we
evaluate performance by comparing the average realized treatment effect and
average predicted treatment effect within bins formed by sorting study
participants according to predicted treatment effect as demonstrated in
\figurename~\ref{fig:lalonde}. Applying the causal forest predictor to the
original features results in a root mean square difference between the average
predicted and realized treatment effects of 4729.51. If the transformed
features are used instead, this discrepancy improves to 3114.82.

From a practical perspective, one may learn the \cate{} in order to select a
subset of people for whom a given intervention has an expected net benefit (and
then deliver that intervention only to persons predicted to benefit from it).
When we focus on the 20\% of people predicted to benefit most from this
treatment, we find that the estimated realized benefit for those chosen using
the transformed features (\$4732.89) is much greater than the benefit for those
chosen using the original feature set (\$816.92). This can be seen visually in
\figurename~\ref{fig:lalonde} by comparing the estimated realized average
treatment effect for bin \#5 (the rightmost bin) in both plots.

\section{Conclusions}
\label{s:conclusions}

Causal inference, especially on real life datasets, poses significant
challenges but offers a crucial avenue for predicting the impact of potential
interventions. Learned feature representations help us to better infer the
conditional average treatment effect, improving our ability to individually
tailor predictions and target subsets of the general population. In this paper,
we propose and validate a novel representation-based method that uses a
neuroevolutionary approach to remove information from features irrelevant for
predicting the outcome. We demonstrate that this method can yield improved
estimates for heterogeneous treatment effects on standard synthetic examples
and illustrate its use on a real life dataset. We believe that representational
learning is particularly well-suited for removing extraneous information in
causal models and we anticipate future research in this area.

\subsubsection*{Acknowledgements.}

We would like to thank Binjie Lai, Yi-Hong Kuo, and Xiang Wu for their support
and feedback. We are also grateful to the anonymous reviewers for their insights
and suggestions.


\appendix

\section{Implementation details}\label{details}

All numerical experiments were performed with Python 3.7.7 and R 3.6.0. We used
Python packages (versioning in parentheses) Keras~(1.0.8), LightGBM~(3.1.1),
Matplotlib~(3.3.4), Numpy~(1.20.2), Pandas~(1.2.4), rpy2~(2.9.4),
Scikit-learn~(0.24.2), Scipy~(1.6.2), Tensorflow~(2.0.0) with Intel MKL
optimizations, and XGBoost~(1.3.3), along with R package grf~(1.2.0).

\end{document}